\documentclass[10pt,twocolumn,letterpaper]{article}

\usepackage{wacv}
\usepackage{times}
\usepackage{epsfig}
\usepackage{graphicx}
\usepackage{amsmath}
\usepackage{amssymb}
\usepackage[dvipsnames]{xcolor}

\usepackage{cite}
\usepackage{subfig}
\usepackage{booktabs}
\usepackage{multirow}
\usepackage{adjustbox}
\usepackage{array}
\usepackage{pifont}% http://ctan.org/pkg/pifont
\newcommand{\cmark}{\ding{51}}%
\newcommand{\xmark}{\ding{55}}%

\newcommand\blfootnote[1]{%
  \begingroup
  \renewcommand\thefootnote{}\footnote{#1}%
  \addtocounter{footnote}{-1}%
  \endgroup
}

\def\wacvPaperID{304} % Enter the WACV Paper ID here

\wacvfinalcopy % *** Uncomment this line for the final submission
\ifwacvfinal
\def\assignedStartPage{1} % *** Enter the assigned starting page number (instead of 9876)
\fi

\ifwacvfinal
\usepackage[breaklinks=true,bookmarks=false]{hyperref}
\else
\usepackage[pagebackref=true,breaklinks=true,colorlinks,bookmarks=false]{hyperref}
\fi

\ifwacvfinal
\else
\fi

\begin{document}

%%%%%%%%% TITLE
\title{Class Anchor Clustering: A Loss for Distance-based Open Set Recognition}

\author{Dimity Miller, Niko S\"underhauf, Michael Milford, Feras Dayoub\\
Queensland University of Technology, Australian Centre for Robotic Vision\\

{\tt\small \{d24.miller, niko.suenderhauf, michael.milford, feras.dayoub\}@qut.edu.au}
}

\maketitle

\blfootnote{The authors acknowledge continued support from the Queensland 
University of Technology (QUT) through the Centre for Robotics. This research was conducted by the Australian Research Council Centre of Excellence for Robotic Vision (project number CE140100016).}

%%%%%%%%% ABSTRACT
\begin{abstract}
In open set recognition, deep neural networks encounter object classes that were unknown during training. Existing open set classifiers distinguish between known and unknown classes by measuring distance in a network's logit space, assuming that known classes cluster closer to the training data than unknown classes. However, this approach is applied post-hoc to networks trained with cross-entropy loss, which does not guarantee this clustering behaviour. To overcome this limitation, we introduce the Class Anchor Clustering (CAC) loss. CAC is a distance-based loss that explicitly trains known classes to form tight clusters around anchored class-dependent centres in the logit space. We show that training with CAC achieves state-of-the-art performance for distance-based open set classifiers on all six standard benchmark datasets, with a 15.2\% AUROC increase on the challenging TinyImageNet, without sacrificing classification accuracy. We also show that our anchored class centres achieve higher open set performance than learnt class centres, particularly on object-based datasets and large numbers of training classes.

\end{abstract}

\newcommand{\vect}[1]{\mathbf{ #1}}
\newcommand{\vectg}[1]{{\boldsymbol{ #1}}}
\newcommand{\ggo}{\ensuremath{\mathrm{g^2o}} }
\newcommand{\R}{\mathbb{R}}
\newcommand{\N}{\mathbb{N}}
\newcommand{\Z}{\mathbb{Z}}
\renewcommand{\P}{\mathbb{P}}
\newcommand{\tran}{^\top}
\newcommand{\T}{^\mathsf{T}}
\newcommand{\iT}{^{-\mathsf{T}}}
\newcommand{\inv}{^{-1}}
\newcommand{\func}[2]{\mathtt{#1}\left\{#2\right\}}
\newcommand{\sig}{\operatorname{sig}}
\newcommand{\diag}{\operatorname{diag}}
\newcommand{\argmin}{\operatornamewithlimits{argmin}}
\newcommand{\argmax}{\operatornamewithlimits{argmax}}
\newcommand{\RMSE}{\operatorname{RMSE}}
\newcommand{\RMSEpos}{\operatorname{RMSE}_\text{pos}}
\newcommand{\RMSEori}{\operatorname{RMSE}_\text{ori}}
\newcommand{\RPE}{\operatorname{RPE}}
\newcommand{\RPEpos}{\operatorname{RPE}_\text{pos}}
\newcommand{\RPEori}{\operatorname{RPE}_\text{ori}}
\newcommand{\rpe}{\varepsilon_{\vdelta}}
\newcommand{\achiError}{\bar{e}_{\chi^2}}
\newcommand{\chiError}{e_{\chi^2}}
\newcommand{\normal}[2]{\mathcal{N}\left(#1, #2\right)}
\newcommand{\uniform}[2]{\mathcal{U}\left(#1, #2\right)}
\newcommand{\pfrac}[2]{\frac{\partial #1}{\partial #2}}  % partielles Differential
\newcommand{\fracpd}[2]{\frac{\partial #1}{\partial #2}} % partielles Differential
\newcommand{\fracppd}[2]{\frac{\partial^2 #1}{\partial #2^2}}  % zweite partielle 
\newcommand{\dd}{\mathrm{d}}  
\newcommand{\smd}[2]{\left\| #1 \right\|^2_{#2}}
\newcommand{\E}[1]{\text{\normalfont{E}}\left[ #1 \right]}     % Expectation operator
\newcommand{\Cov}[1]{\text{\normalfont{Cov}}\left[ #1 \right]} % Covariance operator
\newcommand{\Var}[1]{\text{\normalfont{Var}}\left[ #1 \right]} % Variance operator
\newcommand{\Tr}[1]{\text{\normalfont{tr}}\left( #1 \right)}   % Trace of a matrix
\def\sgn{\mathop{\mathrm sgn}}    
\newcommand{\twovector}[2]{\begin{pmatrix} #1 \\ #2 \end{pmatrix}} % Vektor mit zwei zwei Einträgen
\newcommand{\smalltwovector}[2]{\left(\begin{smallmatrix} #1 \\ #2 \end{smallmatrix}\right)} 
\newcommand{\threevector}[3]{\begin{pmatrix} #1 \\ #2 \\ #3 \end{pmatrix}} % Vektor mit drei Einträgen
\newcommand{\fourvector}[4]{\begin{pmatrix} #1 \\ #2 \\ #3 \\ #4 \end{pmatrix}}  % Vektor mit vier Einträgen
\newcommand{\smallthreevector}[3]{\left(\begin{smallmatrix} #1 \\ #2 \\ #3 \end{smallmatrix}\right)} % kleiner Vektor mit drei Einträgen
\newcommand{\fourmatrix}[4]{\begin{pmatrix} #1 & #2 \\ #3 & #4 \end{pmatrix}} % 2x2-Matrix
\newcommand{\vA}{\vect{A}}
\newcommand{\vB}{\vect{B}}
\newcommand{\vC}{\vect{C}}
\newcommand{\vD}{\vect{D}}
\newcommand{\vE}{\vect{E}}
\newcommand{\vF}{\vect{F}}
\newcommand{\vG}{\vect{G}}
\newcommand{\vH}{\vect{H}}
\newcommand{\vI}{\vect{I}}
\newcommand{\vJ}{\vect{J}}
\newcommand{\vK}{\vect{K}}
\newcommand{\vL}{\vect{L}}
\newcommand{\vM}{\vect{M}}
\newcommand{\vN}{\vect{N}}
\newcommand{\vO}{\vect{O}}
\newcommand{\vP}{\vect{P}}
\newcommand{\vQ}{\vect{Q}}
\newcommand{\vR}{\vect{R}}
\newcommand{\vS}{\vect{S}}
\newcommand{\vT}{\vect{T}}
\newcommand{\vU}{\vect{U}}
\newcommand{\vV}{\vect{V}}
\newcommand{\vW}{\vect{W}}
\newcommand{\vX}{\vect{X}}
\newcommand{\vY}{\vect{Y}}
\newcommand{\vZ}{\vect{Z}}
\newcommand{\va}{\vect{a}}
\newcommand{\vb}{\vect{b}}
\newcommand{\vc}{\vect{c}}
\newcommand{\vd}{\vect{d}}
\newcommand{\ve}{\vect{e}}
\newcommand{\vf}{\vect{f}}
\newcommand{\vg}{\vect{g}}
\newcommand{\vh}{\vect{h}}
\newcommand{\vi}{\vect{i}}
\newcommand{\vj}{\vect{j}}
\newcommand{\vk}{\vect{k}}
\newcommand{\vl}{\vect{l}}
\newcommand{\vm}{\vect{m}}
\newcommand{\vn}{\vect{n}}
\newcommand{\vo}{\vect{o}}
\newcommand{\vp}{\vect{p}}
\newcommand{\vq}{\vect{q}}
\newcommand{\vr}{\vect{r}}
\newcommand{\vt}{\vect{t}}
\newcommand{\vu}{\vect{u}}
\newcommand{\vv}{\vect{v}}
\newcommand{\vw}{\vect{w}}
\newcommand{\vx}{\vect{x}}
\newcommand{\vy}{\vect{y}}
\newcommand{\vz}{\vect{z}}
\newcommand{\valpha}{\vectg{\alpha}}
\newcommand{\vbeta}{\vectg{\beta}}
\newcommand{\vgamma}{\vectg{\gamma}}
\newcommand{\vdelta}{\vectg{\delta}}
\newcommand{\vepsilon}{\vectg{\epsilon}}
\newcommand{\vtau}{\vectg{\tau}}
\newcommand{\vmu}{\vectg{\mu}}
\newcommand{\vphi}{\vectg{\phi}}
\newcommand{\vPhi}{\vectg{\Phi}}
\newcommand{\vpi}{\vectg{\pi}}
\newcommand{\vPi}{\vectg{\Pi}}
\newcommand{\vPsi}{\vectg{\Psi}}
\newcommand{\vchi}{\vectg{\chi}}
\newcommand{\vvarphi}{\vectg{\varphi}}
\newcommand{\veta}{\vectg{\eta}}
\newcommand{\viota}{\vectg{\iota}}
\newcommand{\vkappa}{\vectg{\kappa}}
\newcommand{\vlambda}{\vectg{\lambda}}
\newcommand{\vLambda}{\vectg{\Lambda}}
\newcommand{\vnu}{\vectg{\nu}}
\newcommand{\vgo}{\vectg{\o}}
\newcommand{\vvarpi}{\vectg{\varpi}}
\newcommand{\vtheta}{\vectg{\theta}}
\newcommand{\vvartheta}{\vectg{\vartheta}}
\newcommand{\vrho}{\vectg{\rho}}
\newcommand{\vsigma}{\vectg{\sigma}}
\newcommand{\vSigma}{\vectg{\Sigma}}
\newcommand{\vvarsigma}{\vectg{\varsigma}}
\newcommand{\vupsilon}{\vectg{\upsilon}}
\newcommand{\vomega}{\vectg{\omega}}
\newcommand{\vOmega}{\vectg{\Omega}}
\newcommand{\vxi}{\vectg{\xi}}
\newcommand{\vXi}{\vectg{\Xi}}
\newcommand{\vpsi}{\vectg{\psi}}
\newcommand{\vzeta}{\vectg{\zeta}}
\newcommand{\vzero}{\vect{0}}
\newcommand{\cA}{\mathcal{A}}
\newcommand{\cB}{\mathcal{B}}
\newcommand{\cC}{\mathcal{C}}
\newcommand{\cD}{\mathcal{D}}
\newcommand{\cE}{\mathcal{E}}
\newcommand{\cF}{\mathcal{F}}
\newcommand{\cG}{\mathcal{G}}
\newcommand{\cH}{\mathcal{H}}
\newcommand{\cI}{\mathcal{I}}
\newcommand{\cJ}{\mathcal{J}}
\newcommand{\cK}{\mathcal{K}}
\newcommand{\cL}{\mathcal{L}}
\newcommand{\cM}{\mathcal{M}}
\newcommand{\cN}{\mathcal{N}}
\newcommand{\cO}{\mathcal{O}}
\newcommand{\cP}{\mathcal{P}}
\newcommand{\cQ}{\mathcal{Q}}
\newcommand{\cR}{\mathcal{R}}
\newcommand{\cS}{\mathcal{S}}
\newcommand{\cT}{\mathcal{T}}
\newcommand{\cU}{\mathcal{U}}
\newcommand{\cV}{\mathcal{V}}
\newcommand{\cW}{\mathcal{W}}
\newcommand{\cX}{\mathcal{X}}
\newcommand{\cY}{\mathcal{Y}}
\newcommand{\cZ}{\mathcal{Z}}
\newcommand{\fA}{\mathfrak{A}}
\newcommand{\fB}{\mathfrak{B}}
\newcommand{\fC}{\mathfrak{C}}
\newcommand{\fD}{\mathfrak{D}}
\newcommand{\fE}{\mathfrak{E}}
\newcommand{\fF}{\mathfrak{F}}
\newcommand{\fG}{\mathfrak{G}}
\newcommand{\fH}{\mathfrak{H}}
\newcommand{\fI}{\mathfrak{I}}
\newcommand{\fJ}{\mathfrak{J}}
\newcommand{\fK}{\mathfrak{K}}
\newcommand{\fL}{\mathfrak{L}}
\newcommand{\fM}{\mathfrak{M}}
\newcommand{\fN}{\mathfrak{N}}
\newcommand{\fO}{\mathfrak{O}}
\newcommand{\fP}{\mathfrak{P}}
\newcommand{\fQ}{\mathfrak{Q}}
\newcommand{\fR}{\mathfrak{R}}
\newcommand{\fS}{\mathfrak{S}}
\newcommand{\fT}{\mathfrak{T}}
\newcommand{\fU}{\mathfrak{U}}
\newcommand{\fV}{\mathfrak{V}}
\newcommand{\fW}{\mathfrak{W}}
\newcommand{\fX}{\mathfrak{X}}
\newcommand{\fY}{\mathfrak{Y}}
\newcommand{\fZ}{\mathfrak{Z}}

\section{Introduction}
\label{sec::intro}

Many practical applications require the deployment of trained visual perception models under \emph{open set} conditions, such as autonomous systems, driverless cars, and robotics. In open set conditions, a model encounters object classes that were not present during training (referred to as `unknown' classes)~\cite{scheirer2013toward}. Deep convolutional neural networks (CNNs) degrade in performance in open set conditions, as they can confidently misclassify unknown classes as known training classes~\cite{nguyen2015deep, hendrycks2017baseline, blum2019fishyscapes}. This behaviour raises serious concerns about the safety of using CNNs in open set environments~\cite{amodei2016concrete} -- particularly on autonomous systems where perception failures may have severe consequences~\cite{sunderhauf2018limits,blum2019fishyscapes}. 

\emph{Open set recognition} extends object recognition to an open set environment~\cite{scheirer2013toward}. During testing, an open set classifier must classify known object classes and reject unknown object classes~\cite{scheirer2013toward}. In this paper, we propose a new distance-based loss that achieves state-of-the-art performance for distance-based open set recognition. 

\begin{figure*}[t]
    \centering
    \subfloat[`Ideal' open set recognition]{\includegraphics[width=0.28\linewidth]{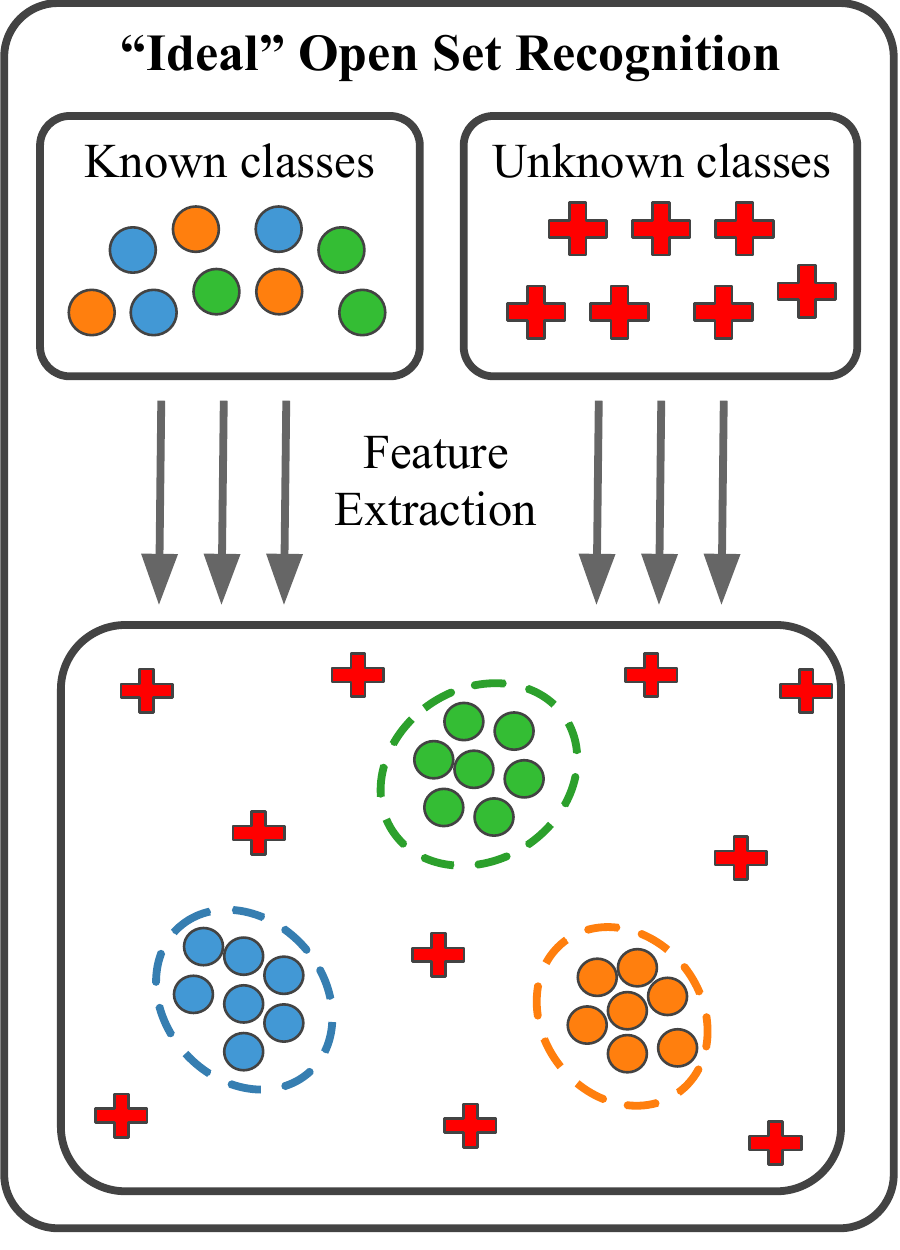} \label{fig:ideal}}
    \subfloat[Open set recognition with cross-entropy loss]{\includegraphics[width=0.35\linewidth]{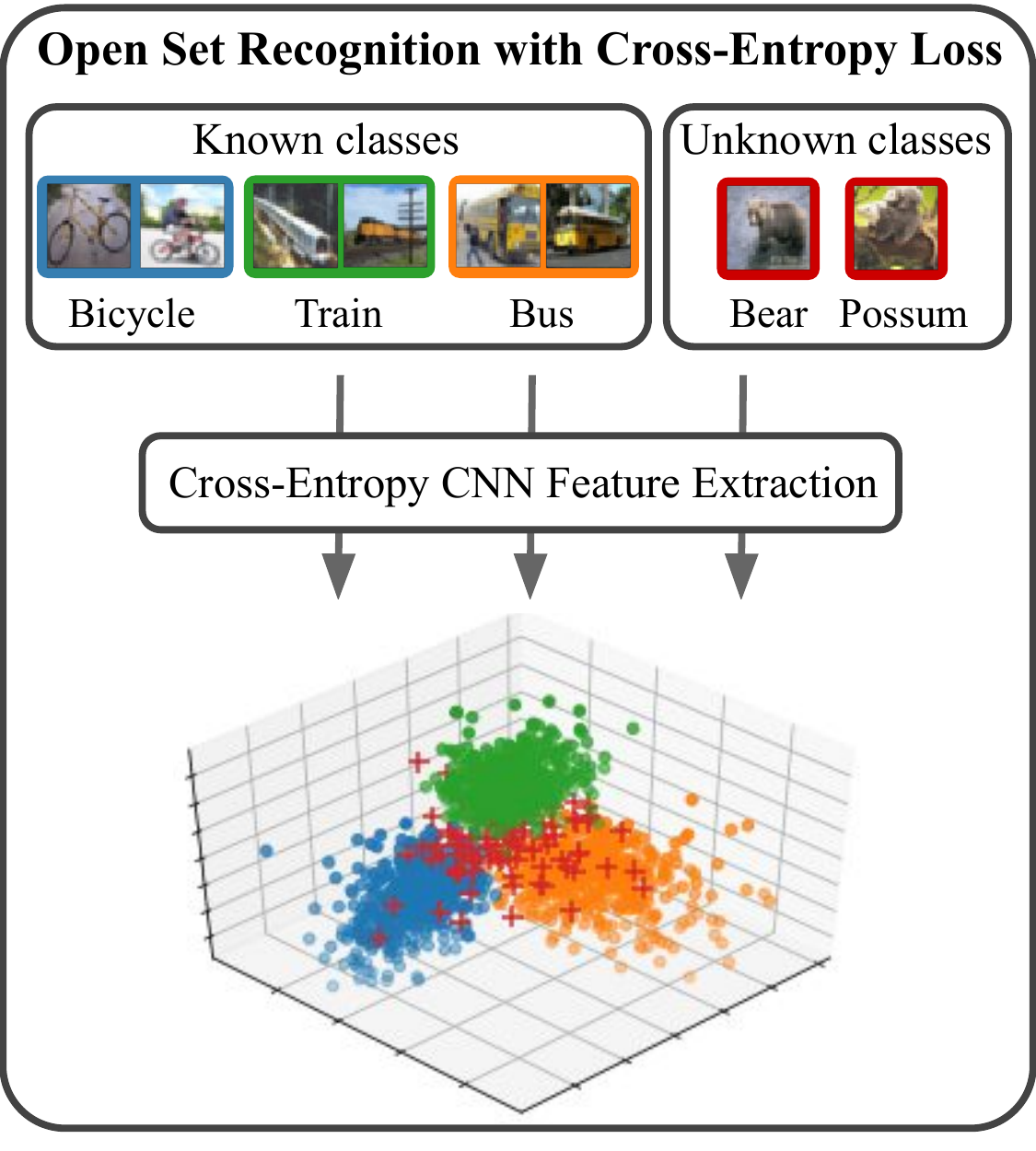} \label{fig:ce}}
    \subfloat[Open set recognition with our CAC loss]{\includegraphics[width=0.35\linewidth]{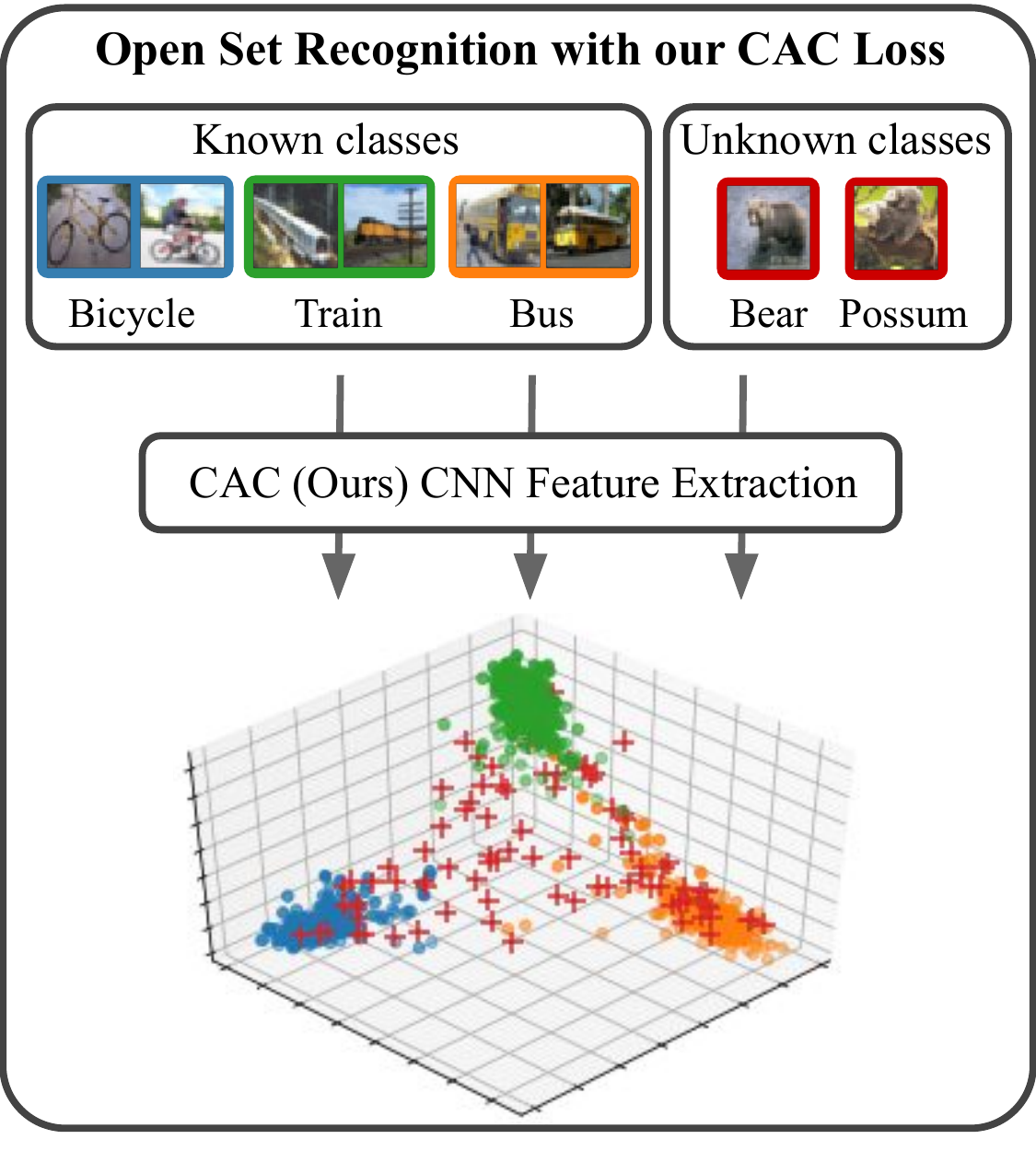} \label{fig:cac}}
    \caption{Left: An `ideal' open set classifier will tightly cluster known classes in the feature space, and unknown classes will fall far away. Middle: A CNN trained on real image data from CIFAR100 with cross-entropy loss shows a final 3-D logit space with inflated known class clusters which cannot be easily distinguished from the unknown classes. Right: A CNN trained with our proposed CAC loss (on the same CIFAR100 data) shows a final 3-D logit space with tight, separated known class clusters and improved distinction of unknown classes.}
    \label{fig:overview}
\end{figure*}

Many open set classifiers model the position of known training data in the final layer, or logit space, of a CNN~\cite{bendale2016towards, yoshihashi2019classification, ge2017generative}. Such approaches assume known classes cluster tightly in the logit space, and that unknown classes will maintain a distance from these clusters. Figure \ref{fig:ideal} shows this ideal performance. Currently, this concept is applied to networks trained with cross-entropy loss~\cite{bendale2016towards, yoshihashi2019classification, ge2017generative}. However, cross-entropy loss does not guarantee the clustering behaviour these methods seek to exploit. We exhibit this in Figure~\ref{fig:ce}, where we train a CNN with cross-entropy loss to classify trains, buses, and bicycles (CIFAR100 classes). The resulting logit space of this CNN appears crowded with inflated class clusters, and it is challenging to distinguish the unknown classes (bear and possum) from these clusters.

In this work, we introduce the Class Anchor Clustering (CAC) loss to address this limitation in prior work. CAC is a distance-based loss that explicitly encourages known training data to form tight clusters around anchored, class-specific centres in the logit space. CAC is compatible with existing classification networks, with only slight modifications to the network architecture. Compared to the cross-entropy trained CNN, a logit space trained with CAC exhibits tight, separate class clusters and an improved distinction of these clusters from unknown classes (see Figure~\ref{fig:cac}). 

Our paper makes the following contributions:
\begin{enumerate}
    \item We propose a new loss term for open set recognition that encourages known class training data to cluster tightly around class-specific centres in logit space. 
    \item We show that training with this novel \emph{Class Anchor Clustering} (CAC) loss achieves new state-of-the-art open set performance for distance-based open set classifiers, without sacrificing classification accuracy.
    \item We introduce the concept of \emph{anchored} class centres as an effective and scalable strategy for distance-based training. In contrast to learning class centres during training, distance losses using anchored centres perform well with object-based datasets with high intra-class variation and large numbers of classes.
\end{enumerate}

\section{Related Work}
\paragraph{Open set recognition}
Open set recognition is multiclass classification with the additional requirement of rejecting inputs from unknown classes \cite{scheirer2013toward}. This is formalised as the task of minimising open space risk, the portion of classification space labelled as `known' that is far from the known training data, while maintaining generalisation and classification accuracy on the known classes~\cite{scheirer2013toward}. Related areas, such as out-of-distribution and novelty detection, exist as relaxed forms of open set classification where known and unknown classes are from different distributions \cite{hendrycks2017baseline} or multiclass classification is not required \cite{boult2019learning}. For this work, we focus specifically on open set recognition.

OpenMax was one of the first CNN open set classifiers, using the network's final layer's logits, or logit space, as the classification space with open space risk~\cite{bendale2016towards}. OpenMax models each known class as a single cluster, and uses a Weibull distribution to re-calibrate softmax scores based on an input's distance to each cluster centre. OpenMax was the first `distance-based' approach, using distance from the training data to minimise the open space risk of a CNN. 

Several following works employed real or generated `known unknown' data to augment the training dataset, either using the data to improve the feature representation for distance-based measures \cite{ge2017generative} or to bound the known classification space with an `other' class \cite{neal2018open, schlachter2019open}. In \cite{dhamija2018reducing}, a network is trained to produce low feature magnitudes and uniform confidence scores for `known unknown' data. While feature magnitudes are used for open set recognition, this method does not model or measure class-specific distances in the logit space \cite{dhamija2018reducing} and therefore we do not define it as a distance-based method.

Other recent open set classifiers use a combined classifier and autoencoder network architecture \cite{yoshihashi2019classification, oza2019c2ae}.  In \cite{oza2019c2ae}, the reconstruction error from a class-conditioned autoencoder-classifier is used to distinguish between known and unknown inputs. Others~\cite{shafaei2018less, denouden2018improving} observed that reconstruction error alone is not suitable as a measure of class novelty. In contrast,  \cite{yoshihashi2019classification} jointly applies OpenMax to a classifier logit space and auto-encoder latent space, with the additional reconstruction-learnt features improving the overall feature representation. Another approach \cite{perera2019deep} uses a self-supervision loss with random transformations to learn a more descriptive feature representation \cite{perera2019deep}. Additionally, the input to the network is augmented with their reconstruction from an autoencoder to further enable open set recognition \cite{perera2019deep}.

In contrast to existing distance-based open set classifiers~\cite{bendale2016towards, yoshihashi2019classification, ge2017generative}, which \emph{assume} known classes will tightly cluster but train with cross-entropy loss, our work is the first to train with a distance loss when using distance in the logit space for testing.

\paragraph{Distance Losses for Deep Neural Networks}
The field of metric learning uses distance loss functions to learn meaningful feature embeddings. Triplet loss is a popular distance loss that encourages inputs to minimise distance to a `positive’ example and maximise distance to a single `negative’ example. Tuplet loss was introduced as an extension of triplet loss that maximises an input’s distance to \emph{multiple} `negative’ examples \cite{sohn2016improved}. We adopt a modified version of Tuplet loss as one of two terms in our new CAC loss but show that Tuplet loss alone is not sufficient for best performance.

Center Loss \cite{wen2016discriminative} was proposed to improve discriminative learning for facial recognition by encouraging clustering in a feature space. It is used in conjunction with cross-entropy loss and encourages an input to minimise distance to its ground truth class centre. The class centres are learnt simultaneously with the feature embedding during training. In contrast, we propose to use \emph{anchored}, i.e. fixed, class centres. This makes training more stable and, as we will show, more scalable to larger and more complex datasets.

Recently, \cite{meyer2019importance} demonstrated the utility of metric learning for open set classification, however only for fine-grained image classification. Such metric learning approaches compute distances between individual instances of the training data, and the sampling technique used can have a significant effect on the convergence speed and stability of the training minimum \cite{wu2017sampling}. As discussed in~\cite{qian2018large}, this sampling typically makes metric learning computationally intractable on larger datasets, such as CIFAR10, CIFAR100, or ImageNet. Although recent work ~\cite{qian2018large} adapted metric learning approaches for large-scale datasets, this technique degraded the classification accuracy of a standard cross-entropy network.

\section{Class Anchor Clustering (CAC) for Open Set Classification}
\label{sec:method}

We now introduce the two core ideas of our paper that enable distance-based training for large-scale image open set classification: (1) the Class Anchor Clustering (CAC) loss that encourages training data to form tight, class-specific clusters. Tight clusters make it easier to distinguish between known and unknown class inputs during deployment.
(2) the concept of using \emph{anchored} class centres in the logit space to fix cluster centre positions during training.

Before introducing our approach, we briefly explain why cross-entropy loss is not sufficient for distance-based open set recognition. Cross-entropy loss minimises the negative log-probability of an input's ground truth class, which is obtained by normalising the logits with a softmax function. The softmax function is \emph{not injective} and multiple input logit vectors map to the same output softmax probability vector \cite{gao2017properties}. As a result, cross-entropy loss cannot guarantee clustering in the logit space.

\paragraph{General Architecture}
CAC is compatible with existing classification networks, with slight modifications to the architecture.
Our proposed CAC-trained open set classifier has three main components: 
\begin{enumerate}
    \item A base network, $\mathit{f}$, that projects an input image \textbf{x} to a vector of class logits $\mathbf{z} = f(\vx)$. This network can be any existing classifier with an N-dimensional logit space, where N is the number of known classes. 
    \item A non-trainable parameter, $\mathbf{C}$, representing a set of class centre points $(\vc_1, \dots, \vc_N)$, one for each of the $N$ known classes.
    \item A new layer, $e(\vz, \vC)$, that calculates $\vd$, a vector of Euclidean distances between a logit vector $\mathbf{z}$ and the set of class centres $\mathbf{C}$.
\end{enumerate}

In summary, the output of our distance-based classifier is 
\begin{equation}
    \vd = e(\vz, \vC) = \left(\|\vz-\vc_1\|_2, \dots, \|\vz-\vc_N\|_2\right)\T
    \label{eq:distvect}
\end{equation}
where $\|\cdot\|_2$ denotes the Euclidean norm.

\subsection{Training with a Distance-based Loss Function}
\label{sec:anchor}
During training, we wish to learn a logit space embedding $ f(\vx) $ where known inputs form tight, class-specific clusters. This clustering enables us to use a distance-to-class-centre metric during testing to reject unknown class inputs and classify known class inputs. 
%To achieve this, we introduce our proposed CAC loss term and the concept of anchored class centres.
\subsubsection{Class Anchor Clustering Loss}
We require a distance-based loss that a) encourages training inputs to minimise the distance to their ground-truth class centre, while b) maximising the distance to all other class centres to encourage discriminative learning. 

To do this, we use a modified Tuplet loss term $\mathcal{L}_{T}$ \cite{sohn2016improved} that forces an input $\mathbf{x}$ to maximise the difference in distance to the correct class centre $\vc_y$ and all other class centres. Remembering that $\vd = (d_1, \dots, d_N)\T$ is defined as in~(\ref{eq:distvect}), we define this loss component as

\begin{equation}
   \mathcal{L}_{T}(\mathbf{x}, y) = \log\Big(1 + \sum_{j \neq y}^{N} e^{d_{y} - d_{j}}\Big).
\end{equation}

$\cL_T$ differs from Tuplet loss~\cite{sohn2016improved} because it is based on class centres $\vC$ rather than sampled class instances. Our modified Tuplet loss term is equivalent to cross-entropy loss applied to the distance vector $\vd$, but used with a $\operatorname{softmin}$ function rather than $\operatorname{softmax}$ (see supplementary material for proof). The $\operatorname{softmin}$ function is the opposite of $\operatorname{softmax}$: it assigns a large value ($\approx 1$) to the \emph{smallest} value of the input vector and is defined as:
\begin{equation}
    \operatorname{softmin}({\mathbf{d}})_{i} = \frac{e^{-{\mathbf{d}}_{i}}}{\sum_{k = 1}^{N}e^{-{\mathbf{d}}_{k}}}.
\end{equation}

While effective for discriminative learning, $\cL_T$ aims to maximise the \emph{margin} between distance to the correct class centre and distance to the incorrect class centres. To ensure an input is explicitly forced to lower its \emph{absolute} distance to the correct class centre, we also penalise the Euclidean distance between the training logit and the ground truth class centre. We refer to this as the Anchor loss term:

\begin{equation}
    \mathcal{L}_{A}(\mathbf{x}, y) = d_y =  \| \mathit{f}(\mathbf{x}) - \mathbf{c}_{y} \|_2.
\end{equation}

We combine the Anchor and Tuplet loss terms to form our final distance-based loss, which we refer to as the Class Anchor Clustering (CAC) loss:

\begin{equation}
    \mathcal{L}_\text{CAC}(\vx, y) = \mathcal{L}_T(\vx, y) + \lambda \mathcal{L}_A(\vx, y).
\end{equation}

A hyperparameter of our method is $\lambda$, which balances these two individual loss terms (explored in section \ref{sec::hyperparameter}). By combining the Anchor and Tuplet loss terms, our loss minimises training inputs distance to their ground-truth anchored class centre, while maximising the distance to other anchored class centres.

\subsubsection{Anchored Class Centres}
We introduce anchored class centre points as a method of \emph{anchoring}, i.e. fixing, cluster centres for each class in the logit space during training. By anchoring our class centres during training, we eliminate the need to learn another parameter (as done in previous approaches to distance losses, e.g.~\cite{wen2016discriminative}). 

For each known class $i$, our network has an anchored class centre $\mathbf{c}_i$ in the logit space. Given an $N$-dimensional logit space for $N$ known classes, we place the anchored centre for each known class at a point on its class coordinate axis. This anchored centre point is therefore equivalent to a scaled standard basis vector $\ve_i$, or scaled one-hot vector, for each class. The magnitude of the anchored centre, $\alpha$, is a hyperparameter of our method (explored in section \ref{sec::hyperparameter}). We summarise this below:

\begin{align}
    \vC = (\vc_1, \dots, \vc_N) = (\alpha\cdot\ve_1, \dots, \alpha\cdot\ve_N) \\
    \ve_1 = (1,0,\dots,0)\T, \;\;\; \ve_N = (0,\dots,0,1)\T.
\end{align}

After completing training, the anchored class centre positions \textbf{C} are adjusted to the mean position of the correctly classified training data. This allows us to model the class cluster centres for complex datasets more accurately, where visual and semantic similarities between classes can cause slight divergence from the original anchored class centre positions. Note that while anchoring our class centres equal distances apart in the logit space may limit the learning of semantically meaningful features, for this work we aim to learn only a discriminative feature representation that exhibits tight clustering behaviour.

\subsection{Using Distance-based Measures during Testing}
During testing, the network has to reject unknown class inputs and correctly classify known class inputs. Our CAC loss trains known inputs to have two distance-based properties: (1) a high $\operatorname{softmin}$ score for the distance to correct known class centre (as per the modified Tuplet loss term $\cL_T$) and (2) a low absolute distance to the correct known class centre (as per the Anchor loss term $\cL_A$). Based on this, we calculate rejection scores $\vgamma = (\gamma_1, ..., \gamma_N)\T$ that express the classifier's \emph{dis}belief that the input $\vx$ belongs to each of the $N$ known classes. We calculate the rejection scores $\vgamma$ as the element-wise product ($\circ$) of the distance vector $\vd$ and its inverted $\operatorname{softmin}$:
\begin{equation}
    \label{eq:gamma}
    \vgamma = \vd \circ (1 - \operatorname{softmin}(\vd))
\end{equation}

By weighting the absolute distance with the inverted $\operatorname{softmin}$ score, inputs must have both a low absolute distance and high $\operatorname{softmin}$ score to be assigned a low rejection score for a known class. If all values in $\vgamma$ are above a threshold $\theta$, the input does not belong to any known class and is rejected as unknown. Otherwise, the class label corresponding to the smallest value in $\vgamma$ is assigned:
\begin{equation}
    \text{decision} = 
    \begin{cases}
        \text{rejected as unknown} & \text{if } \min(\vgamma) > \theta \\
        \text{class } i = \argmin{\vgamma} & \text{if } \min(\vgamma) \le \theta
    \end{cases}
    \label{eq:decision}
\end{equation}

Using this distance-based decision procedure minimises open space risk~\cite{scheirer2013toward}: the further away an input $\vx$ projects from the class-specific centres, the more likely it is to be rejected as unknown.

\section{Experimental Setup} \label{sec::setup}
To simulate open set conditions, a set of `known' and `unknown' classes must be established. The \emph{openness} $\vO$ of the classification task \cite{scheirer2013toward} can then be defined as
\begin{equation}
    \mathbf{O} = 1 - \sqrt{\frac{2 \cdot N_\text{train}}{N_\text{test} + N_\text{target}}}
\end{equation}
\noindent
where $N_\text{train}$ is the number of `known' classes during training, $N_\text{target}$ is the number of `known' classes for classification during testing and $N_\text{test}$ is the total number of testing classes (`known' and `unknown'). A higher openness indicates a more difficult problem setup, but other factors such as the visual similarity between known and unknown classes also influence the difficulty. We follow the established benchmark evaluation protocol \cite{neal2018open}, where standard classification datasets are adapted to the open set task by randomly splitting into `known’ and `unknown’ classes.

\subsection{Datasets} 
The details of each dataset in its open set configuration are summarised below. For each dataset, performance is evaluated over 5 trials with random known and unknown class splits.  
\\
\textbf{MNIST} \cite{lecun2010mnist}: grayscale $32\times32$ images of handwritten digits, 6 known and 4 unknown classes, \textbf{O} = 13.39\%.\\
\textbf{SVHN} \cite{netzer2011reading}: RGB $32\times32$ images of street view house digits, 6 known and 4 unknown classes, \textbf{O} = 13.39\%.  \\
\textbf{CIFAR10} \cite{krizhevsky2009learning}: RGB $32\times32$ images of animals and objects, 6 known and 4 unknown classes, \textbf{O} = 13.39\%. \\
\textbf{CIFAR+10/+50}: considers the 4 non-animal classes of CIFAR10 as known, and 10 or 50 randomly sampled animal classes from CIFAR100~\cite{krizhevsky2009learning} as unknown (\textbf{O} = 33.33\% and 62.86\%). \\
\textbf{TinyImageNet} \cite{web:tinyimagenet}: RGB $64\times64$ images of animals and objects, 20 known and 180 unknown classes, \textbf{O} = 57.35\%. TinyImageNet images can contain significant background information unrelated to the object class, a number of classes are very visually and semantically related (e.g. different breeds of dogs), and there is high visual variation within individual classes. Examples are provided in the supplementary material.

\subsection{Metrics}
We use the following metrics to assess the performance of an open set classifier.\\
\noindent
\textbf{Area Under the ROC Curve (AUROC)} is a calibration-free measure of the open set performance of a classifier. The Receiver Operating Characteristic (ROC) curve represents the trade-off between true positive rate (known inputs correctly retained as `known') and the false positive rate (unknown inputs incorrectly retained as `known') when applying varying thresholds to a given score. We modify the threshold $\theta$ that is compared to our network's rejection scores $\vgamma$ as discussed in~(\ref{eq:decision}).\\
\noindent
\textbf{Classification Accuracy} measures the classifier's accuracy when applied to only the known classes in the dataset, equivalent to closed set classification. An open set classifier should maintain the classification accuracy of a standard closed set classifier.\\
\noindent
\textbf{Correct Classification Rate (CCR)} measures the fraction of known inputs that are correctly retained as `known' \emph{and} correctly classified as their ground truth class \cite{dhamija2018reducing}. It can be computed for various false positive rates when varying a threshold (we threshold our network's rejection scores $\vgamma$).

\subsection{State-of-the-art Methods for Comparison}
We compare to seven existing state-of-the-art open set classifiers~\cite{hendrycks2017baseline, neal2018open, oza2019c2ae, perera2020generative, bendale2016towards, ge2017generative, yoshihashi2019classification}, following the established protocol that uses AUROC as the primary evaluation metric. The results of these experiments are shown in Table \ref{tab:main_results}, where we also highlight the three methods~\cite{yoshihashi2019classification,ge2017generative,bendale2016towards} that use distance in the logit space during \emph{testing} to distinguish between known and unknown inputs. In contrast to CAC, none of those methods uses distance during \emph{training}.

We additionally compare to another state-of-the-art open set classifier \cite{dhamija2018reducing} that uses a different evaluation protocol (as described in Section \ref{sec::resultsSOTA}).

Instead of assessing the overall performance with AUROC, \cite{dhamija2018reducing} evaluate with CCR, which quantifies performance at specific false positive operating points. This is an important metric for many safety-critical domains where performance at reasonable low false positive rates is relevant. We hope that including this comparison encourages future open set works to do so also.

\subsection{Implementation Details}
We use the network architecture specified by the benchmark evaluation protocol \cite{neal2018open}. We use a Stochastic Gradient Descent (SGD) optimizer with a learning rate of 0.01 and train until convergence. We then complete another training cycle with a lower learning rate of 0.001 and train again until convergence. More details about the training procedure are in the supplementary material. For all datasets, we use an Anchor loss weight $\lambda$ of 0.1 and a logit anchor magnitude $\alpha$ of 10.

\begin{table*}[t!]
    % \footnotesize
    \setlength{\tabcolsep}{4pt}
    \centering
    \begin{tabular}{@{}lccccccccc@{}}
        \toprule
        \textbf{Method} & \multicolumn{2}{c}{\textbf{Use Distance in}} & \textbf{MNIST} & \textbf{SVHN} & \textbf{CIFAR10} & \textbf{CIFAR+10} &\textbf{CIFAR+50} & \textbf{TinyImageNet} \\
        & \textbf{Training} & \textbf{Testing} & & & & & \\
        \midrule
        Softmax\cite{hendrycks2017baseline} & \textcolor{red}{\xmark} & \textcolor{red}{\xmark} & 97.8 $\pm$ 0.6 & 88.6 $\pm$ 1.4 & 67.7 $\pm$ 3.8 & 81.6 $\pm$ N.R. & 80.5 $\pm$ N.R. & 57.7 $\pm$ N.R. \\
        OSRCI\cite{neal2018open} & \textcolor{red}{\xmark} & \textcolor{red}{\xmark} & \textit{98.8 $\pm$ 0.4} & 91.0 $\pm$ 1.0 & 69.9 $\pm$ 3.8 & 83.8 $\pm$ N.R. & 82.7 $\pm$ N.R. & 58.6 $\pm$ N.R. \\
        C2AE\cite{oza2019c2ae} & \textcolor{red}{\xmark} & \textcolor{red}{\xmark} & - & 89.2 $\pm$ 1.3 & 71.1 $\pm$ 0.8 & 81.0 $\pm$ 0.5 & 80.3 $\pm$ 0.0 & 58.1 $\pm$ 1.9\\
        GFROR \cite{perera2020generative} &\textcolor{red}{\xmark} & \textcolor{red}{\xmark} & - & \textit{93.5 $\pm$ 1.8} & \textbf{80.7 $\pm$ 3.9} & \textbf{92.8 $\pm$ 0.2} & \textbf{92.6 $\pm$ 0.0} & \textit{60.8 $\pm$ 1.7} \\
        OpenMax\cite{bendale2016towards} & \textcolor{red}{\xmark} & \textcolor{OliveGreen}{\cmark} & 98.1 $\pm$ 0.5 & 89.4 $\pm$ 1.3 & 69.5 $\pm$ 4.4 & 81.7 $\pm$ N.R. & 79.6 $\pm$ N.R. & 57.6 $\pm$ N.R.\\
        G-OpenMax\cite{ge2017generative} & \textcolor{red}{\xmark} & \textcolor{OliveGreen}{\cmark} & 98.4 $\pm$ 0.5 & 89.6 $\pm$ 1.7 & 67.5 $\pm$ 4.4 & 82.7 $\pm$ N.R. & 81.9 $\pm$ N.R. & 58.0 $\pm$ N.R. \\
        CROSR\cite{yoshihashi2019classification} & \textcolor{red}{\xmark} & \textcolor{OliveGreen}{\cmark} & \textbf{99.1 $\pm$ 0.4} & 89.9 $\pm$ 1.8 & - & - & - & 58.9 $\pm$ N.R.\\
    
        \midrule
  
        \textbf{CAC (Ours)} & \textcolor{OliveGreen}{\cmark} & \textcolor{OliveGreen}{\cmark} & \textbf{99.1 $\pm$ 0.5} & \textbf{94.1 $\pm$ 0.7} & \textit{80.1 $\pm$ 3.0} & \textit{87.7 $\pm$ 1.2} & \textit{87.0 $\pm$ 0.0} & \textbf{76.0 $\pm$ 1.5}\\
        \bottomrule
    \end{tabular}
    \caption{Open set AUROC for state-of-the-art methods and our proposed approach. Best and second best performance are bolded and italicised respectively.\vspace{-0.3cm}} 
    \label{tab:main_results}
\end{table*}

\section{Results and Discussion}

Our evaluation revealed four main results that we discuss in the following: 
(1) CAC outperforms the existing distance-based open set classifiers~\cite{yoshihashi2019classification,ge2017generative,bendale2016towards} on every tested dataset, without sacrificing classification accuracy (Section \ref{sec::resultsSOTA}). 
(2) Compared to other distance losses, CAC achieves better open set performance (Section \ref{sec::resultsOthers}).
(3) Training with \emph{anchored} class centres achieves better open set performance than learnt class centres on nearly all tested datasets, particularly on object-based datasets with high intra-class visual variations. Anchored centres also maintain open set performance better with increasing numbers of known classes (Section \ref{sec::resultsAnchors}).
(4) Training with CAC is insensitive to the choice of its two hyperparameters over a wide range of values (Section \ref{sec::hyperparameter}).

% ===========================================================================
\subsection{Comparison with State-of-the-Art Open Set Classifiers}
\label{sec::resultsSOTA}
The open set performance of our proposed approach is compared to the state-of-the-art methods in Table \ref{tab:main_results}.

\paragraph{Comparison with other distance-based approaches:}  Compared to other state-of-the-art methods that use distance in the logit space during testing~\cite{bendale2016towards, ge2017generative, yoshihashi2019classification}, we achieve the best open set performance on all six of the benchmark datasets. Our performance increase is most substantial on TinyImageNet and CIFAR10, where there is an increase of 17.1\% and 10.6\%.

\begin{figure}[tb]
    \centering
    \includegraphics[width=.47\textwidth]{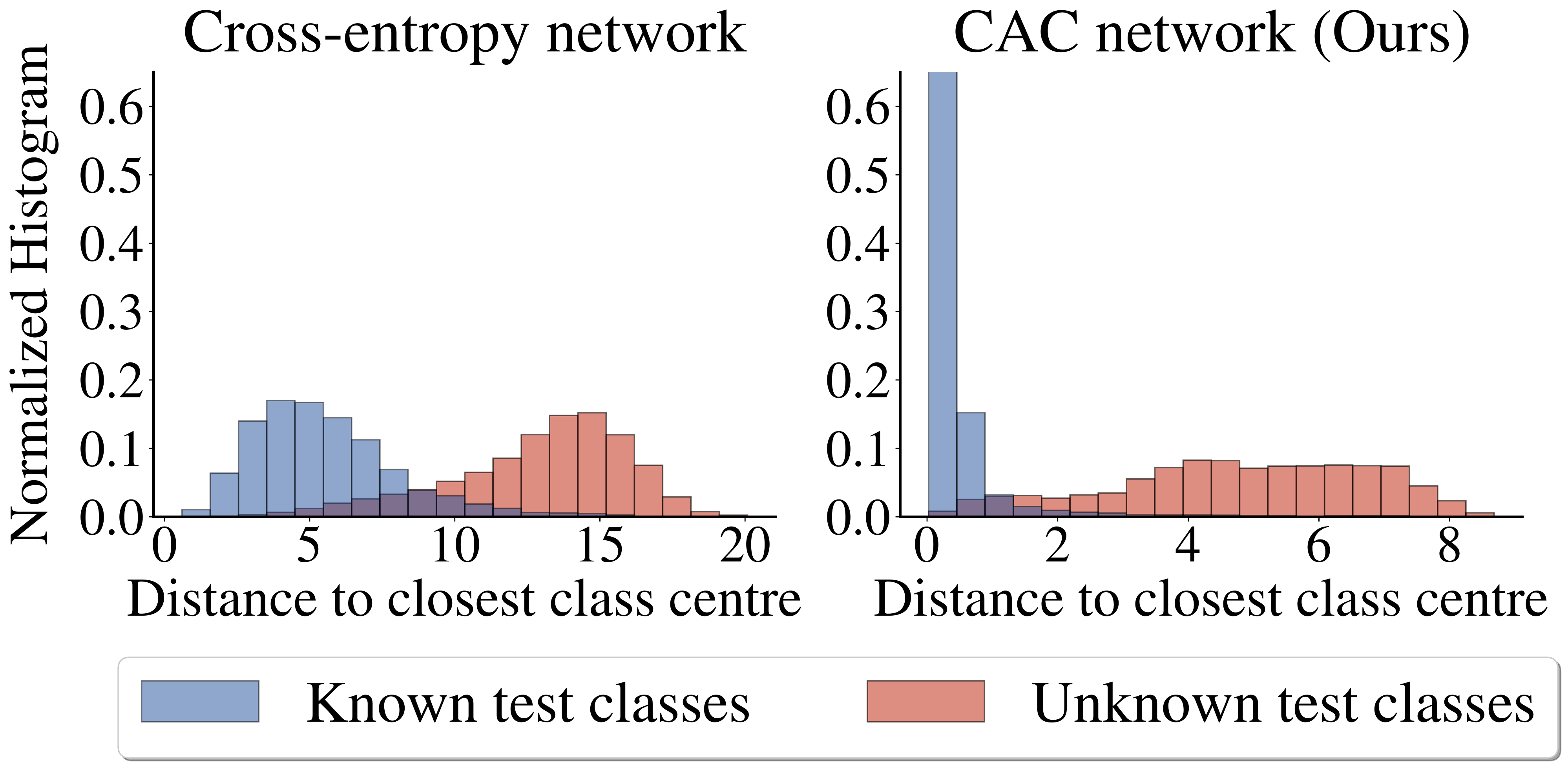}
    \caption{Training with CAC (right) causes known test data to cluster more tightly to the class centres than training with cross-entropy. This allows for distance to better separate known and unknown test data. This plot shows the distributions for MNIST, see Table \ref{tab:cluster} for other datasets.}
    
    \label{fig:histo}
\end{figure}

\begin{table}[tb]
    \setlength{\tabcolsep}{3pt}
    \centering
    \begin{tabular}{@{}lcc@{}}
        \toprule
        \textbf{Dataset} & \textbf{Cross-Entropy} & \textbf{CAC (Ours)} \\
        \midrule
        MNIST & 0.414 & \textbf{0.324}\\
        SVHN & 0.700 & \textbf{0.573}\\
        CIFAR10 & 0.946 & \textbf{0.868}\\
        CIFAR+10 & 0.899 & \textbf{0.766}\\
        CIFAR+50 & 0.889 & \textbf{0.751}\\
        TinyImageNet & 0.984 & \textbf{0.913}\\
        
        \bottomrule
    \end{tabular}
    \caption{Compared to cross-entropy loss, training with CAC gives a lower Bhattacharyya coefficient between distributions of known and unknown class distances to the closest class centre. This represents less overlap between distributions, enabling better distance-based open set recognition.}
    \label{tab:cluster}
\end{table}

Our proposed approach is the first method that trains with a distance-based loss when using distance in the logit space during testing. To analyse the impact of distance-based training, we examine the distributions of known class and unknown class distances to a class centre for a network trained with cross-entropy loss (as used by ~\cite{bendale2016towards, ge2017generative, yoshihashi2019classification}) and a network trained with our proposed CAC loss. As shown in Figure \ref{fig:histo}, the CAC-trained network has a known distribution that clusters more tightly to the class centres (behaviour it was trained for), and as a result, there is a lower overlap with the unknown distribution. By reducing the overlap with the unknown distribution, the open set classifier can more accurately identify and reject unknown inputs, thus improving open set performance.

In Table \ref{tab:cluster}, we quantitatively show that training with CAC decreases the overlap between the known and unknown class distance distributions in comparison to cross-entropy loss. The table shows the Bhattacharyya coefficient~\cite{bhattacharyya1943measure}, an established measure of the overlap between two distributions. For each of the datasets, CAC loss results in a lower Bhattacharyya coefficient, on average by 14.3\%.

\paragraph{Comparison to non-distance-based approaches:}
Compared to non-distance-based open set classifiers, we achieve state-of-the-art performance on TinyImageNet, MNIST and SVHN, and come second to GFROR~\cite{perera2020generative} on CIFAR10 and CIFAR+10/+50.

While CAC achieves a 15.2\% performance increase on TinyImageNet with 20 known classes, it performs less well to \cite{perera2020generative} on CIFAR+10/+50 variations with only 4 known classes. When presented with only 4 known classes, CAC has less data to learn a rich feature representation that ensures known and unknown class inputs do not project to the same region in the logit space. In contrast, \cite{perera2020generative} specifically uses reconstruction and self-supervision techniques during training to improve the feature representation.

\begin{figure}[tb]
    \centering
    \includegraphics[width=.47\textwidth]{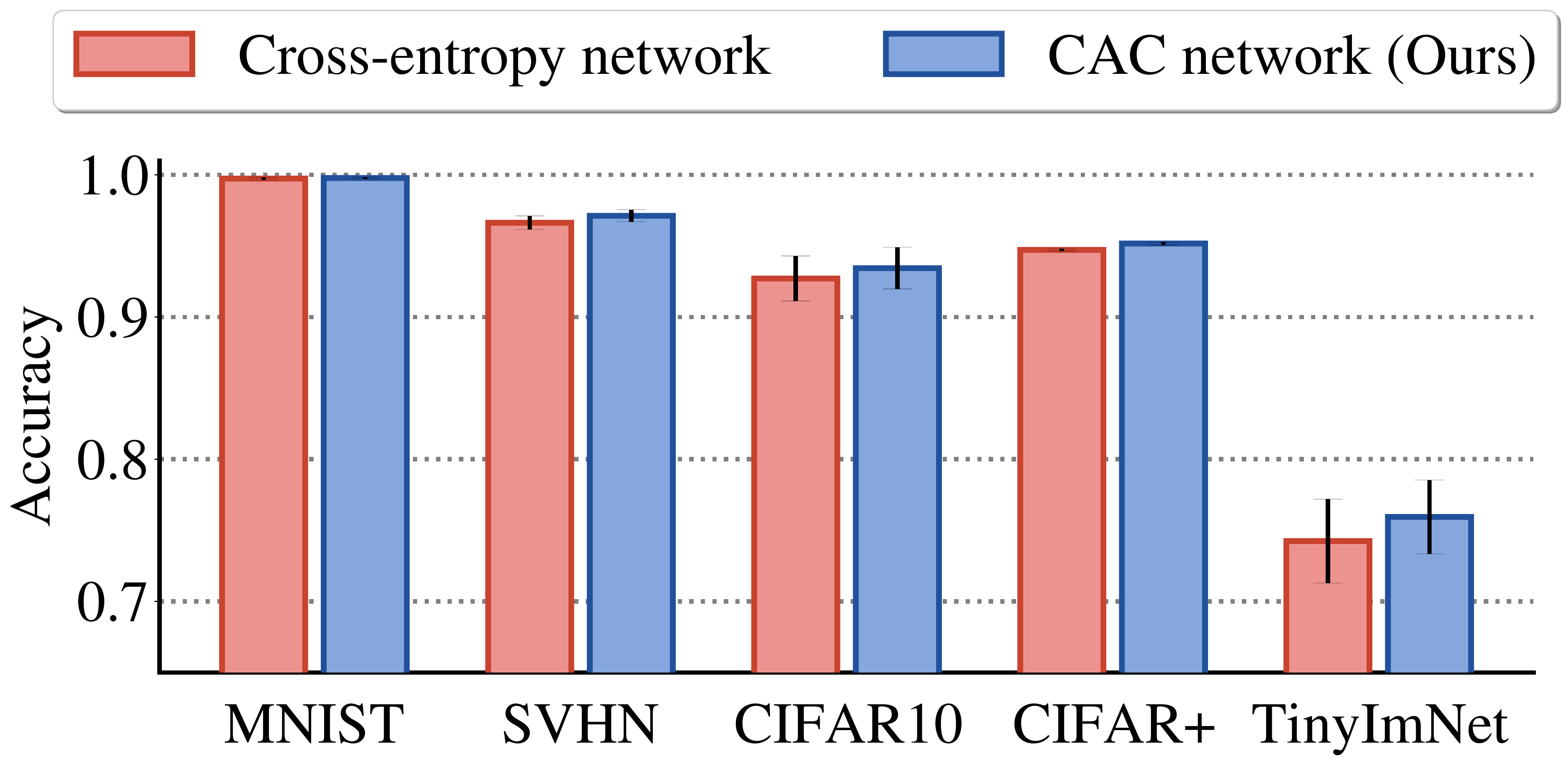}
    \vspace{-0.2cm}
    \caption{Our CAC classifier maintains the classification accuracy of a standard classifier trained with cross-entropy loss.}
    \label{fig:acc}
\end{figure}

\paragraph{Maintaining classification accuracy:} In Figure~\ref{fig:acc}, we show that training with CAC loss maintains the closed set classification accuracy of a standard network. The standard network uses the same architecture but is trained with cross-entropy loss and uses the softmax score for classification. This result demonstrates that our method improves open set performance without compromising classification accuracy.

\begin{table}[tb]
    % \footnotesize
    \setlength{\tabcolsep}{4pt}
    \centering
    \begin{tabular}{@{}ccccc@{}}
        \toprule
        \textbf{Unknown Data} & \textbf{Method} & \multicolumn{3}{c}{\textbf{CCR at FPR of}}\\
        & & \textbf{1\%} & \textbf{5\%} & \textbf{10\%} \\
        \midrule
        \multirow{3}{*}{SVHN} & Softmax \cite{hendrycks2017baseline} & 46.0 & N.R. & 64.7 \\
        & Objectosphere \cite{dhamija2018reducing}$^\dagger$ & 54.5 & N.R. & 70.1 \\
        & CAC (Ours) & \textbf{68.3} & \textbf{78.5} & \textbf{82.6} \\
        \midrule
        \multirow{3}{*}{\shortstack{CIFAR100\\ Subset}} & Softmax \cite{hendrycks2017baseline} & 23.4 & N.R. & 51.4 \\
        & Objectosphere \cite{dhamija2018reducing}$^\dagger$ & \textbf{43.3} & N.R. & 66.5 \\
        & CAC (Ours) & 32.5 & \textbf{63.8} & \textbf{73.8} \\
        \bottomrule
    \end{tabular}
    \caption{At False Positive Rates (FPR) 1\%, 5\% and 10\%, CAC achieves state-of-the-art Correct Classification Rates (CCR) for the unknown datasets (with CIFAR10 as the known dataset). $^\dagger$\cite{dhamija2018reducing} uses a separate subset of CIFAR100 as known unknowns during training.
    %Best performance bolded. 
    \vspace{-0.3cm}} 
    \label{tab:ccr_results}
\end{table}

\paragraph{Open set performance at low false positive operating points:} 
For many safety-critical domains (e.g. robotics or medical applications), the performance at specific low false positive rates is practically relevant. Although the main benchmark used for open set classification (in Table~\ref{tab:main_results}) does not provide such an evaluation, we can compare CAC's performance at different false positive rates against~\cite{dhamija2018reducing} and~\cite{hendrycks2017baseline}. Table~\ref{tab:ccr_results} shows the results after training a ResNet-18 on CIFAR10, and using a CIFAR100 subset and SVHN as the unknown datasets. We note that~\cite{dhamija2018reducing} uses a `known unknown' subset of CIFAR100 during training, which potentially contributes to their better performance at the very low 1\% FPR on the CIFAR100 experiment. We refer the reader to the Supplementary for details on the experimental setup.  

% ===========================================================================

\subsection{Ablation Studies}
% ===========================================================================
\subsubsection{Comparison with Existing Distance Losses}
\label{sec::resultsOthers}
We proposed CAC loss specifically for the task of \emph{training} a distance-based open set classifier. However, other distance losses have been proposed for other computer vision tasks, e.g. metric learning~\cite{sohn2016improved} and facial recognition~\cite{wen2016discriminative}. In this experiment, we compare the open set performance achieved when training with Center loss~\cite{wen2016discriminative}, Tuplet loss~\cite{sohn2016improved}, the Anchor loss component $\cL_A$ of CAC, and our proposed CAC loss. We train the same network architecture with each loss function and use our anchored class centres. Table \ref{tab:otherLosses} summarises the open set AUROC results for each of the losses.

CAC outperforms all other distance losses~\cite{wen2016discriminative, sohn2016improved} on SVHN, CIFAR10, CIFAR+10, CIFAR+50 and TinyImageNet, with Center loss~\cite{wen2016discriminative} achieving second best performance. Center loss uses cross-entropy loss on the logits to \emph{implicitly} encourage inputs to maximise distance to other class centres. In contrast, CAC \emph{explicitly} forces this behaviour by applying Tuplet loss directly to the output distance vector. Interestingly, when used alone, our Anchor loss term and Tuplet loss cannot achieve the same performance as when they are combined to create CAC loss. This validates that both loss terms are important for distance-based open set classification, as together they simultaneously ensure minimised distance to the correct class centre as well as maximised distance to all other class centres. 

\begin{table}[tb]
    \setlength{\tabcolsep}{4pt}
    \centering
    \begin{tabular}{@{}lcccc@{}}
        \toprule
        \textbf{Dataset} & \textbf{Center} & \textbf{Tuplet} & \textbf{$\cL_A$ only} & \textbf{CAC} \\
        & \textbf{\cite{wen2016discriminative}} & \textbf{\cite{sohn2016improved}} & \textbf{(Ours)} & \textbf{(Ours)} \\
        \midrule
        MNIST & \textbf{0.988} & 0.957 & 0.979& 0.987  \\
        SVHN & 0.941 & 0.833 & 0.888 & \textbf{0.942} \\
        CIFAR10 & 0.786 & 0.739 & 0.751 & \textbf{0.803} \\
        CIFAR+10 & 0.854 & 0.844 & 0.804 & \textbf{0.863} \\
        CIFAR+50 & 0.863 & 0.837 & 0.816 & \textbf{0.872} \\
        TinyImageNet & 0.765 & 0.717 & 0.749 & \textbf{0.772} \\
        
        \bottomrule
    \end{tabular}
    \caption{CAC provides better open set AUROC performance than the compared distance losses on nearly all the benchmark datasets. \vspace{-0.3cm}}
    \label{tab:otherLosses}
\end{table}

\subsubsection{Anchored versus Learnt Class Centres}
\label{sec::resultsAnchors}
In this section we investigate the benefits of using \emph{anchored} class centres in the context of open set classification. While our work is the first to anchor class centres during the training process, 
previous distance losses such as Center loss~\cite{wen2016discriminative} encourage clustering around class centres that are simultaneously \emph{learnt} during training.

We compare the open set performance when training with learnt and anchored class centres, and repeat this experiment with CAC loss and Center loss~\cite{wen2016discriminative}. To learn class centres, we use the approach described in \cite{wen2016discriminative}. Learning centres with CAC required the addition of cross-entropy loss for stability (see supplementary material for details). 

In Table \ref{tab:dynamic}, we show that anchored class centres yield better open set performance than learnt class centres, for both Center loss~\cite{wen2016discriminative} and our proposed CAC loss. The performance difference between anchored and learnt centres is greatest for the object-based datasets (CIFAR10 variants and TinyImageNet), with an average 2.2\% improvement for Center loss~\cite{wen2016discriminative} and 0.85\% for CAC. Learning class centres during training relies on a stable learning signal from the images in each batch. However, CIFAR10 and TinyImageNet can exhibit considerable visual variations within each class, thus providing a potentially noisy learning signal for the class centre positions. By anchoring our class centres in the logit space, we eliminate this difficulty and allow for high performance on object-based datasets.

\begin{table}[tb]
    \setlength{\tabcolsep}{4pt}
    \centering
    \begin{tabular}{@{}lcccc@{}}
        \toprule
        \textbf{Dataset} & \multicolumn{2}{c}{\textbf{Center \cite{wen2016discriminative}}} & \multicolumn{2}{c}{\textbf{CAC (Ours)}} \\
        & \textbf{Learnt} & \textbf{Anchored}  & \textbf{Learnt} & \textbf{Anchored} \\
        \midrule
        MNIST & 0.985 & \textbf{0.988} & 0.987 & \textbf{0.987}  \\
        SVHN & 0.937 & \textbf{0.941} & \textbf{0.946} & 0.942 \\
        CIFAR10 & 0.763 & \textbf{0.786} & 0.791 & \textbf{0.803} \\
        CIFAR+10 & 0.831 & \textbf{0.854} & 0.856 & \textbf{0.863} \\
        CIFAR+50 & 0.848 & \textbf{0.863} & 0.865 & \textbf{0.872} \\
        TinyImNet & 0.738 & \textbf{0.765} & 0.764 & \textbf{0.772} \\
        
        \bottomrule
    \end{tabular}
    \caption{Anchored class centres yields better open set AUROC than learnt class centres, particularly on the object-based CIFAR10 and TinyImageNet datasets.\vspace{-1em}}
    \label{tab:dynamic}
\end{table}

\begin{figure}[tb]
    \centering
    \includegraphics[width=.49\textwidth]{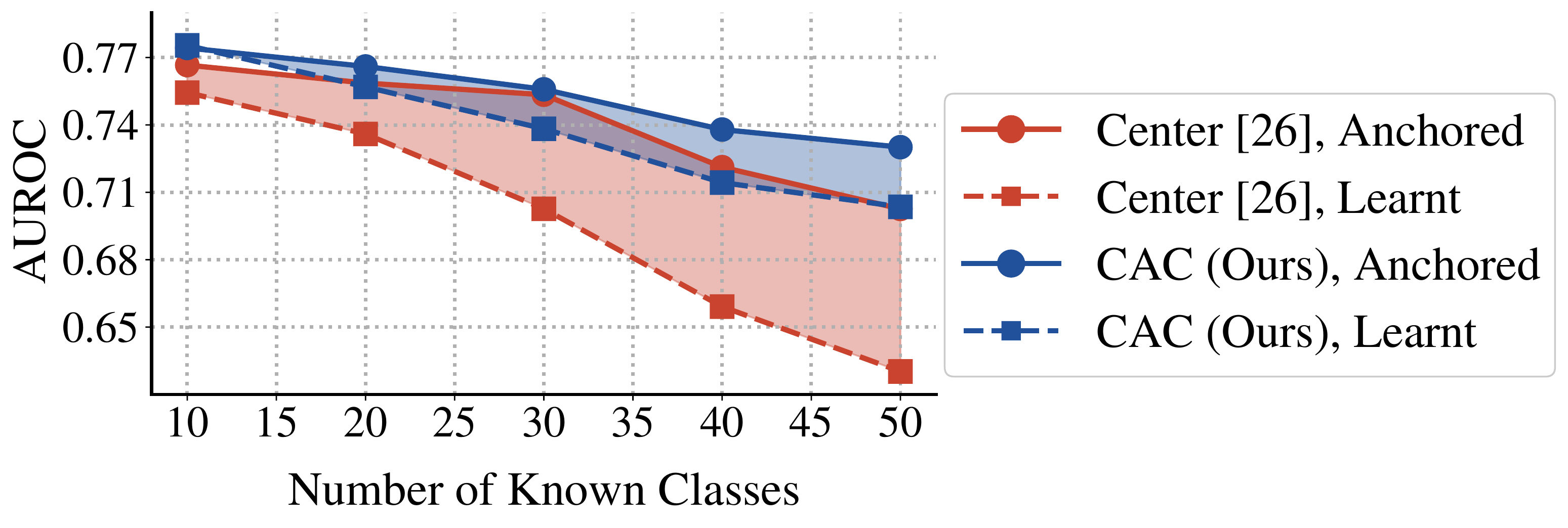}
    \caption{Anchored class centres perform better with increasing numbers of classes than learnt class centres. Results are averaged over 5 trials of random class splits. }
    \label{fig:center}
\end{figure}

Learning class centres is even more difficult for tasks with large numbers of classes, as each batch will provide less data per class. To investigate this effect, we train with learnt and anchored class centres for Center loss~\cite{wen2016discriminative} and CAC loss on \emph{increasing} numbers of known TinyImageNet classes, while keeping the openness of the open set task fixed at 18.35\%. As we can see in Figure~\ref{fig:center}, open set performance of a network trained with learnt class centres degrades at a faster rate than a network trained with anchored class centres for both Center loss~\cite{wen2016discriminative} and CAC loss. 

In summary, we show that training with anchored class centres yields better performance on object-based datasets and scales better to larger numbers of training classes. We found this to be consistent for both tested loss functions. In addition, we observed training with anchored centres requires approximately half the epochs to converge, speeding up the training process (see supplementary material).

\subsubsection{Analysis of Hyperparameters of CAC loss}
\label{sec::hyperparameter}
\begin{figure}[tb]
    \centering
    \includegraphics[width=.49\textwidth]{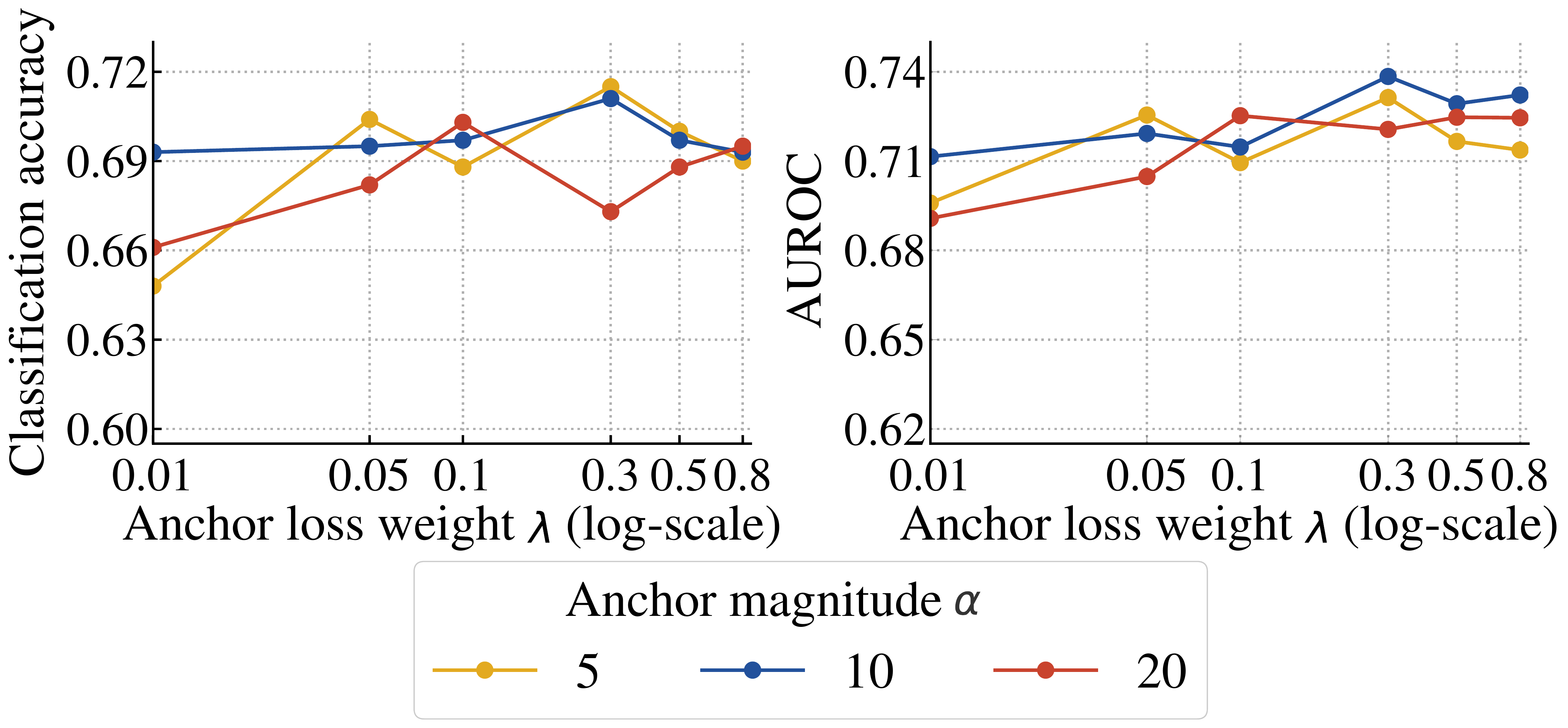}

    \caption{CAC Performance is not sensitive to hyperparameters anchor loss weight $\lambda$ and anchor magnitude $\alpha$ across a range of values (results from 1 trial on TinyImageNet).}
    \label{fig:hyperparameter}
\end{figure}

Our proposed CAC loss has two hyperparameters, the Anchor loss term weight $\alpha$ and the anchored centre magnitude $\lambda$, and their sensitivity is shown in Figure \ref{fig:hyperparameter}. With an Anchor loss weight $0.05 \leq \lambda \leq 0.8$ and an anchor magnitude $5\leq \alpha \leq 20$, both the classification accuracy and open set AUROC vary by less than 4\%.

\section{Conclusions}
The deployment of deep neural networks under open set conditions remains an important and difficult challenge for computer vision. Reliability and robustness in the presence of unknown class inputs is crucial for many safety-critical applications such as driverless cars or robotics.

We introduced and demonstrated the benefits of anchored class centres and the novel Class Anchor Clustering loss for open set recognition. Future work could focus on learning more meaningful and diverse feature representations, potentially by allowing for more complex arrangements of the anchored class centres. A more semantically meaningful feature representation may allow for better distinction of unknown classes and increased interpretability of open set errors, both of which have the potential to improve distance-based open set recognition even further. We additionally look forward to engaging with the community to develop new evaluation protocols and datasets beyond the simple ones commonly used in open set recognition, such as MNIST, SVHN, or even CIFAR10. Many practical applications rely on open set robustness, and we believe benchmark datasets should better reflect the complexity and richness of those real world applications.

\newpage
{\small
\bibliographystyle{ieee_fullname}
\bibliography{refs}
}

\end{document}